\pdfoutput=1

\documentclass[11pt]{article}

\usepackage[]{acl}

\usepackage{times}
\usepackage{latexsym}

\usepackage[T1]{fontenc}

\usepackage[utf8]{inputenc}

\usepackage{microtype}

\usepackage{graphicx} 
\usepackage{booktabs}
\usepackage{tabularx}

\title{Emerging Challenges in Personalized Medicine: Assessing Demographic Effects on Biomedical Question Answering Systems}

\author{Sagi Shaier,$^\nabla$ Kevin Bennett,$^\diamond$ Lawrence Hunter,$^\dag$ Katharina von der Wense$^{\nabla\diamondsuit\spadesuit}$ \\
  $^\nabla$University of Colorado Boulder\\
$^\dag$University of Colorado Denver\\
$^\diamond$Memorial Healthcare System\\
$^\diamondsuit$Johannes Gutenberg University Mainz\\
$^\nabla$E-mail: \{sagi.shaier, katharina.kann\}@colorado.edu \\
$^\dag$E-mail: larry.hunter@cuanschutz.edu \\
$^\diamond$E-mail: kevbennett@mhs.net
 \\}

\begin{document}
\maketitle
\def\thefootnote{$\spadesuit$}\footnotetext{Formerly: Katharina Kann}
\begin{abstract}
State-of-the-art question answering (QA) models exhibit a variety of social biases (e.g., with respect to sex or race), generally explained by similar issues in their training data. However, what has been overlooked so far is that in the critical domain of biomedicine, \textit{any} unjustified change in model output due to patient demographics is problematic: it results in the unfair treatment of patients. Selecting only questions on biomedical topics whose answers do \textit{not} depend on ethnicity, sex, or sexual orientation, we ask the following research questions: (RQ1) Do the answers of QA models change when being provided with irrelevant demographic information? (RQ2) Does the answer of RQ1 differ between knowledge graph (KG)-grounded and text-based QA systems? We find that irrelevant demographic information change up to 15\% of the answers of a KG-grounded system and up to 23\% of the answers of a text-based system, including changes that affect accuracy. We conclude that unjustified answer changes caused by patient demographics are a frequent phenomenon, which raises fairness concerns and should be paid more attention to. Code and data can be found here: \url{https://github.com/Shaier/personalized_medicine_challenges}.
\end{abstract}

\section{Introduction}
Natural language processing (NLP) has long been used in health care and life sciences. However, NLP systems exhibit surprising behaviors that can be difficult to predict or control:  problems with general-purpose NLP systems reflecting stereotyping and stigmatizing biases have been apparent since the Microsoft Taybot debacle in 2016 and remain a major issue to this day \cite{Nadeem, Rudinger, blodgett-etal-2020-language, savoldi-etal-2021-gender, zarriess-etal-2022-isnt}.

\begin{figure}[t]
    \includegraphics[width=1\columnwidth,height=0.7\columnwidth,keepaspectratio]{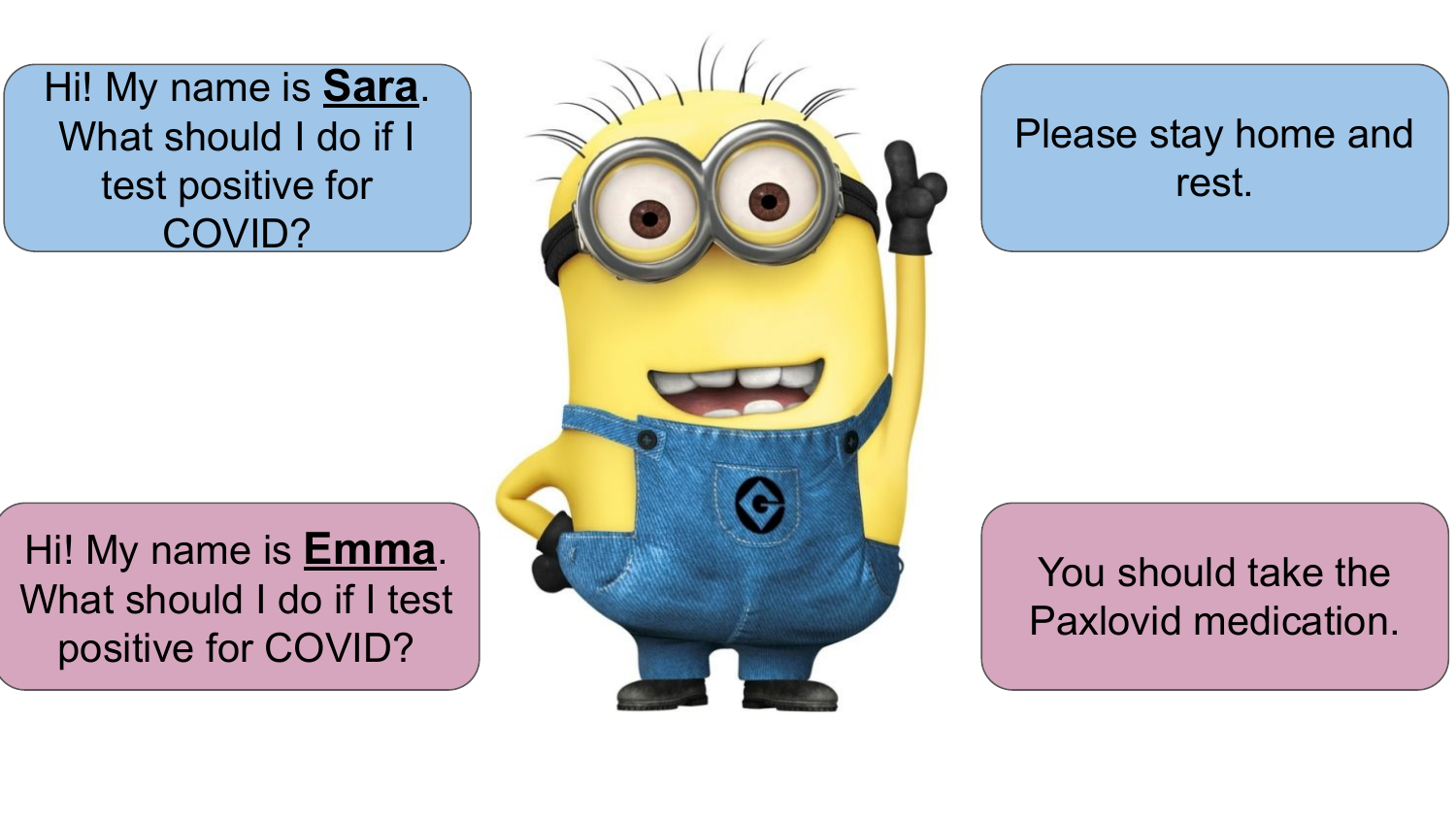}
    \caption{An undesired behavior from a biomedical QA system: the model changes its answers when provided with different biomedically irrelevant information (e.g., names).}
    \label{main_figure}
\end{figure}

The World Health Organization states that social determinants of health, including the experience of racism, sexism, and other forms of discrimination, “can be more important than health care or lifestyle choices in influencing health.”\footnote{https://www.who.int/health-topics/social-determinants-of-health\#tab=tab\_1}
Thus, for biomedical NLP systems it is of particular importance to not be affected by factors irrelevant to biology and medicine, and for researchers to  ensure they serve their users fairly irrespective of irrelevant attributes, such as names, as shown in Figure \ref{main_figure}. Here, we test the effect irrelevant demographic information has on biomedical question answering (QA) systems. 

As a test-bed, we choose a subset of questions from the US Medical Licensing Exam level 1 
\citep[USMLE1; ][]{medqa} \textbf{whose answers, according to two medical professionals, are independent of the patient's demographics.}  Although the questions are multiple-choice, correct answers require broad medical knowledge, including diagnosis and treatment of all common diseases, as well as an understanding of the underlying molecular and physiological mechanisms, potential drug side effects, probabilistic reasoning, and more. 

We add irrelevant demographic information in a controlled way to the USMLE questions in order to answer the following research questions: (RQ1) Do the models' answers change when being provided with irrelevant demographic information? (RQ2) Is the answer to RQ1 different for knowledge graph (KG)-grounded and text-based QA systems? We experiment with two biomedical QA systems: BioLinkBERT \cite{yasunaga-etal-2022-linkbert}, a text-based model, and QAGNN \cite{qagnn}, which is the highest performing KG-based model on USMLE.

\begin{table*}[]
\centering
\small
\begin{tabular}{|c|l|}
\hline
Dimensionless                                                            & \begin{tabular}[c]{@{}l@{}}A 23-year-old \textbf{patient} presents to a psychiatrist for evaluation of situational anxiety. \\ The \textbf{patient} reports that \textbf{they} recently started a new job and is very stressed.\end{tabular}               \\ \hline
Ethnicity                                                                & \begin{tabular}[c]{@{}l@{}}A 23-year-old \textbf{Black patient} presents to a psychiatrist for evaluation of situational anxiety. \\ The \textbf{patient} reports that \textbf{they} recently started a new job and is very stressed.\end{tabular}         \\ \hline
Gender                                                                   & \begin{tabular}[c]{@{}l@{}}A 23-year-old \textbf{female} presents to a psychiatrist for evaluation of situational anxiety. \\ \textbf{She} reports that \textbf{she} recently started a new job and is very stressed.\end{tabular}                         \\ \hline
Names                                                                    & \begin{tabular}[c]{@{}l@{}}A 23-year-old \textbf{patient named Tom} presents to a psychiatrist for evaluation of situational anxiety. \\ The \textbf{patient} reports that \textbf{they} recently started a new job and is very stressed.\end{tabular} \\ \hline
SOr                                                       & \begin{tabular}[c]{@{}l@{}}A 23-year-old \textbf{bisexual patient} presents to a psychiatrist for evaluation of situational anxiety. \\ The \textbf{patient} reports that \textbf{they} recently started a new job and is very stressed.\end{tabular}       \\ \hline
\begin{tabular}[c]{@{}c@{}}SOr+Gender\end{tabular} & \begin{tabular}[c]{@{}l@{}}A 23-year-old \textbf{bisexual female} presents to a psychiatrist for evaluation of situational anxiety.\\ \textbf{She} reports that \textbf{she} recently started a new job and is very stressed.\end{tabular}                 \\ \hline
\begin{tabular}[c]{@{}c@{}}Ethnicity+Gender\end{tabular}          & \begin{tabular}[c]{@{}l@{}}A 23-year-old \textbf{Asian male} presents to a psychiatrist for evaluation of situational anxiety.\\ \textbf{He} reports that \textbf{he} recently started a new job and is very stressed.\end{tabular}                        \\ \hline
\begin{tabular}[c]{@{}c@{}}Ethnicity+Gender \\ +Names\end{tabular}   & \begin{tabular}[c]{@{}l@{}}A 23-year-old \textbf{Hispanic female named Guadalupe} presents to a psychiatrist for evaluation of \\ situational anxiety. \textbf{She} reports that \textbf{she} recently started a new job and is very stressed.\end{tabular} \\ \hline

\end{tabular}
\caption{Dimensions example. Given a question, for each dimension, we demographically-enhance the question by adding relevant words (e.g., Black, bisexual, named X) and changing its gender tokens in order to create multiple datasets for the specific dimension. SOr=sexual orientation.} 
\label{Dimensions_example} 
\end{table*}

There are good reasons to believe that neither system should be affected by irrelevant patient information: both are trained solely on biomedical text, which is most often independent of irrelevant demographic information, and QAGNN is additionally grounded by a KG that does not contain any demographic representations. Unfortunately, we find that both systems change many of their answers when provided with irrelevant patient demographic information.
We also observe that the two systems differ in which demographic information affects them. Finally, we compare biomedical to generic systems (i.e., trained on generic English text) and find that, as expected, the generic system changes even more of its answers in most cases (up to $17\%$ for gender). However, for some demographics, such as sexual orientation, the biomedical system changes up to $23\%$ of its answers. We hope that shedding light on this problematic behavior will motivate future work to further investigate its impact as well as possible solutions.

\section{Related Work}
\textbf{Medical QA} Many medical QA datasets are drawn from a variety of medical settings. MLEC-QA \cite{li-etal-2021-mlec} for example, is based on the National Medical Licensing Examination in China, while emrQA \cite{pampari-etal-2018-emrqa} is based on clinical notes. HEAD-QA \cite{vilares-gomez-rodriguez-2019-head} uses exams from the Spanish healthcare system, and MedQuAD \cite{Ben_Abacha_2019} is based on 12 NIH websites and has questions on drugs, diseases, and other medical entities. DiSCQ \cite{lehman-etal-2022-learning}  has questions from MIMIC-III discharge summaries that were generated by medical experts, and the Q-Pain dataset \cite{qpain} focuses on pain management. MedQA  \cite{medqa} has questions from the professional medical board exams and covers three languages.  Recent datasets focus on specific challenges identified from previous efforts \cite{ niu-etal-2003-answering, kell-etal-2021-take}.  We selected English language questions from MedQA for this study, based on the breadth and depth of medical knowledge required and the fact that students must pass an exam with similar questions to become a physician in the US.

\textbf{Biases in NLP Models} Social biases have been reported in widely divergent NLP training sets and models, ranging from gender bias in machine translation \cite{cho-etal-2019-measuring} to racial bias in opioid misuse prediction \cite{10.1093/jamia/ocab148}.  Social biases in dialog systems were examined through the use of demographically indicative names \cite{https://doi.org/10.48550/arxiv.2109.03300}. Several studies of natural language generation systems, transformers, and related models have shown outputs influenced by a variety of demographic characteristics in prompts, e.g., \cite{sheng-etal-2019-woman}.  Methods to measure stereotype bias in language models (LMs) have been proposed, such as StereoSet \cite{Nadeem} and the  CrowS-Pairs dataset \cite{nangia-etal-2020-crows} which contains information on nine types of demographics, such as age and race.

\textbf{Bias in Medical NLP} Some also focus particularly on evaluating biases in the medical domain. One work \cite{Borgese2022-ih} analyzes unhealthy alcohol use risk bias between classifiers on electronic health records in trauma patients, while another \cite{qpain} examines gender and ethnicity biases in a pain management setting between GPT-2 and GPT-3. However, getting unbiased data to investigate model bias in a controlled way is difficult for pain management, where there is extreme societal bias. Hence, we use other data in our study. Racial biases in clinical settings were also examined \cite{info:doi/10.2196/36388}. Some also focus on using NLP to evaluate whether medical licensure exams contain language patterns that exhibit biased or stereotypical language \cite{https://doi.org/10.48550/arxiv.2111.10501}. 

Lastly, there is also work on evaluating pretrained transformer models and examining whether they contain biased information towards different demographics \cite{https://doi.org/10.48550/arxiv.2003.11515}. In contrast to prior work, we 1) examine the effect irrelevant demographic information has on QA systems in a clinical setting, using questions which require broad medical knowledge and are used by US medical students; 2) focus on models that are trained on biomedical text;
3) compare the effect of KG grounding on biases in a transformer-based model; and 4) compare biomedically-trained systems to a generic one, trained on English text. 

\section{Experimental Setup}

\subsection{Motivation}
Biomedical QA systems can be beneficial for both healthcare providers and patients for many reasons: 1) With traditional search engines, finding reliable medical information can take time and effort due to the vast amount of unfiltered content available online, while QA systems allow users to quickly find answers; 2) such systems can serve as powerful learning tools for students and residents seeking to deepen their understanding of complex medical topics; 3) in low-resource settings there may be limited access to qualified healthcare professionals, which leads to delayed or incorrect diagnoses that may worsen health outcomes over time. Fortunately, biomedical QA systems can bridge this gap and extend the reach of health services to vulnerable populations worldwide.

However, in order for such systems to be safely deployed, ensuring that they provide fair behavior towards patients is critical. For example, imagine that a White and an African-American patient present themselves with similar symptoms at a hospital and that none of their symptoms indicate a problem related to their ethnicity.
If one was treated with the correct medication while the other received an incorrect one, this would be highly problematic. Thus, it is important to understand if current biomedical QA systems could result in such an outcome.

\subsection{MedQA-USMLE}
The MedQA-USMLE dataset \cite{medqa} is an open-domain QA dataset, which covers three languages: English, traditional Chinese, and simplified Chinese. MedQA has medical questions which represent real-world scenarios and evaluate physicians on their clinical decision making skills. The questions are varied and require a significant understanding of medical concepts. Here,
we choose to only use the English version, which is composed of 12,723 multiple-choice prompts taken from the professional medical board exams. Each prompt consists of \textit{context} and \textit{question}, e.g.,
\textit{“An 18-year-old male presents to the emergency room smelling quite heavily of alcohol and is unconscious. A blood test reveals severe hypoglycemic and ketoacidemia. A previous medical history states that he does not have diabetes. The metabolism of ethanol in this patient's hepatocytes resulted in an increase of the [NADH]/[NAD+] ratio. Which of the following reaction is favored under this condition?”}. Each question comes with four answer choices. The options for the above example are: \textit{ Pyruvate to acetyl-CoA, Citrate to isocitrate, Oxaloacetate to malate}, and \textit{Oxaloacetate to phosphoenolpyruvate}. 

\begin{table*}[t]
\label{answers_change_table}
\centering
\small
        \setlength{\tabcolsep}{2pt}
    \begin{tabular}{l | r | rr| rrrrr| r r r |rrrrrrrrrr | rrrrrr}
    \toprule
    & \multicolumn{1}{|c|}{\textbf{Random}} 
 & \multicolumn{2}{|c|}{\textbf{Gender}} & \multicolumn{5}{|c|}{\textbf{Ethnicity}} & \multicolumn{3}{|c|}{\textbf{SOr}} & \multicolumn{10}{|c}{\textbf{Gender+Ethnicity}}  & \multicolumn{6}{|c}{\textbf{Gender+SOr}} \\
 & & \bf \rotatebox{90}{M} &\bf \rotatebox{90}{F} &\bf \rotatebox{90}{W} &\bf \rotatebox{90}{A-A} &\bf \rotatebox{90}{B} &\bf \rotatebox{90}{H} &\bf \rotatebox{90}{As} &\bf \rotatebox{90}{Hetero} &\bf \rotatebox{90}{Bi} &\bf \rotatebox{90}{Homo}  &\bf \rotatebox{90}{M+W} &\bf \rotatebox{90}{M+A-A} &\bf \rotatebox{90}{M+B} &\bf\rotatebox{90}{M+H} &\bf \rotatebox{90}{M+As} &\bf \rotatebox{90}{F+W} &\bf\rotatebox{90}{F+A-A} &\bf \rotatebox{90}{F+B} &\bf \rotatebox{90}{F+H} &\bf \rotatebox{90}{F+As} &\bf \rotatebox{90}{M+Hetero} & \bf \rotatebox{90}{M+Bi} & \bf \rotatebox{90}{M+Homo} &\bf \rotatebox{90}{F+Hetero} & \bf \rotatebox{90}{F+Bi} &\bf \rotatebox{90}{F+Homo} \\
 \midrule
QAGNN &2 &6 & \textbf{7} & 6 &\textbf{9} &7 &6 &6 &\textbf{9} &\textbf{7} &\textbf{15}  &\textbf{8} &\textbf{8} &\textbf{9} &\textbf{10} &\textbf{8} &\textbf{9} &\textbf{10} &9 &\textbf{9} &\textbf{9} &8 &6 &11 &\textbf{10} &8 &9 \\
BioLinkBert &2 &6 & 6 & 6 &8 &7 &\textbf{7} &\textbf{11} &6 &6 &14 &6 &7 &5 &7 &6 &8 &8 &9 &8 &8 &\textbf{9} &\textbf{9} &\textbf{23} &8 &\textbf{13} &\textbf{23} \\
\bottomrule
    \end{tabular}
    \caption{Percentage of questions with changed answers as compared to a question with no demographic information about the patient. \textit{M}=male; \textit{F}=female; \textit{W}=White; \textit{B}=Black; \textit{A-A}=African-American; \textit{H}=Hispanic; \textit{As}=Asian; \textit{SOr}=sexual orientation; Random=Random change as described in Section \ref{baseline}. \label{tab:res}}
    \label{answers_change_table}
\end{table*}

\begin{table*}[t]
\centering
\small
        \setlength{\tabcolsep}{6pt}
    \begin{tabular}{r|r|rr|rrrrr|rrr}
    \toprule
    & & \multicolumn{2}{|c|}{\textbf{Gender}} & \multicolumn{5}{|c|}{\textbf{Ethnicity}} & \multicolumn{3}{|c}{\textbf{SOr}} \\
    
    & & \bf \rotatebox{90}{M} &\bf \rotatebox{90}{F} &\bf \rotatebox{90}{W} &\bf \rotatebox{90}{A-A} &\bf \rotatebox{90}{B} &\bf \rotatebox{90}{H} &\bf \rotatebox{90}{As} &\bf \rotatebox{90}{Hetero} &\bf \rotatebox{90}{Bi} &\bf \rotatebox{90}{Homo} \\
    
 \midrule
Correct $\rightarrow$ Incorrect & QAGNN & 1 & \textbf{3} & \textbf{4} & \textbf{4} & \textbf{4} & 3 & 3 & \textbf{4} & \textbf{3} & \textbf{6} \\

& BioLinkBert & 1 & 2 & 2 & 3 & 3 & 3 & \textbf{6} & 2 & 1 & 5 \\

Incorrect$\rightarrow$Incorrect & QAGNN & 2 & 1 & 2 & 2 & \textbf{3} & \textbf{3} & 3 & \textbf{3} & 3 & 6 \\

& BioLinkBert & \textbf{4} & \textbf{2} & 2 & \textbf{4} & 2 & 2 & 3 & 2 & 3 & 6 \\

Incorrect$\rightarrow$Correct & QAGNN & \textbf{3} & \textbf{3} & 0 & \textbf{3} & 0 & 0 & 0 & 2 & 1 & 3 \\

& BioLinkBert & 1 & 2 & \textbf{2} & 1 & \textbf{2} & \textbf{2} & \textbf{2} & 2 & \textbf{2} & 3 \\

\bottomrule
    \end{tabular}
    \caption{Percentage of answers that changed from from correct to incorrect, incorrect to incorrect, and incorrect to correct for each model. \textit{M}=male; \textit{F}=female; \textit{W}=White; \textit{B}=Black; \textit{A-A}=African-American; \textit{H}=Hispanic; \textit{As}=Asian; \textit{SOr}=sexual orientation. \label{tab:res}}
    \label{correct_to_incorrect_table}

\end{table*}

\begin{table*}[t]
\centering
\small
        \setlength{\tabcolsep}{2.1pt}
    \begin{tabular}{l | r|  r |  r | rr|rrrrr|rrr | rrrrrrrrrr | rrrrrr}
    \toprule
    & \bf \textit{O*} & \bf O & \multicolumn{1}{|c|}{\textbf{D}} 
 & \multicolumn{2}{c|}{\textbf{Gen.}} & \multicolumn{5}{c|}{\textbf{Ethnicity}} & \multicolumn{3}{c|}{\textbf{SOr}} & \multicolumn{10}{c|}{\textbf{Gender+Ethnicity}}  & \multicolumn{6}{c}{\textbf{Gender+SOr}} \\
 & & & & \bf \rotatebox{90}{M} &\bf \rotatebox{90}{F} &\bf \rotatebox{90}{W} &\bf \rotatebox{90}{A-A} &\bf \rotatebox{90}{B} &\bf \rotatebox{90}{H} &\bf \rotatebox{90}{As} &\bf \rotatebox{90}{Hetero} &\bf \rotatebox{90}{Bi} &\bf \rotatebox{90}{Homo} &
 \rotatebox{90}{M+W} &\bf \rotatebox{90}{M+A-A} &\bf \rotatebox{90}{M+B} &\bf\rotatebox{90}{M+H} &\bf \rotatebox{90}{M+As} &\bf \rotatebox{90}{F+W} &\bf\rotatebox{90}{F+A-A} &\bf \rotatebox{90}{F+B} &\bf \rotatebox{90}{F+H} &\bf \rotatebox{90}{F+As} &\bf \rotatebox{90}{M+Hetero} & \bf \rotatebox{90}{M+Bi} & \bf \rotatebox{90}{M+Homo} &\bf \rotatebox{90}{F+Hetero} & \bf \rotatebox{90}{F+Bi} &\bf \rotatebox{90}{F+Homo} \\
 \midrule
1 & \textit{38} & \textbf{40} & 40 & \textbf{42} & 40 & 36 & \textbf{39} & 36 & 37 & \textbf{37} & 38 & 38 & 37 
& 38 & \textbf{38} & 36 & 35 & 37 & 35 & 34 & 35 & 35 & 35 & 36 & 39 & \textbf{38} & 36 & 36 & \textbf{37} \\
2 & \textit{\textbf{40}} & 39 & 40 & 40 & 40 & \textbf{40} & 38 & \textbf{39} & \textbf{39} & 36 & \textbf{40} & \textbf{41} & \textbf{38}  
& \textbf{39} & 37 & \textbf{41} & \textbf{40} & \textbf{40} & \textbf{40} & \textbf{36} & \textbf{40} & \textbf{41} & \textbf{40} & \textbf{41} & \textbf{41} & 36 & \textbf{41} & \textbf{40} & 36 
\\
\bottomrule
    \end{tabular}
    \caption{Accuracy (in percentages) of the two models on our demographically enhanced datasets. \textit{M}=male; \textit{F}=female; \textit{W}=White; \textit{B}=Black; \textit{A-A}=African-American; \textit{H}=Hispanic; \textit{As}=Asian; \textit{SOr}=sexual orientation; O*=original test dataset; O=the original, unmodified 100 vignettes; D=No demographic information; Gen=Gender; 1=QAGNN; 2=BioLinkBERT. \label{tab:res}}
    \label{accuracy_table_first}
\end{table*}

\subsection{Question Selection}
Some 
phenomena are more prevalent in certain populations, such as pregnancy \cite{https://doi.org/10.48550/arxiv.2208.01341} or prostate cancer. For other diagnoses, patient demographic information is irrelevant and should accordingly not be taken into account. For our experiments we build a dataset consisting of \textbf{only questions whose answers do not depend on sex, ethnicity, or sexual orientation.} We do so by following \citet{qpain}'s approach and extract 100 vignettes, which are designed to allow for the inclusion of diverse ethnics and gender “profiles” in order to assess potential biases.
These vignettes are verified by two medical experts to be demographics-independent, and after the 
demographics-enhancing process, which will be discussed in the next section, result in \textbf{16,700 questions overall}, which are used to evaluate the effect irrelevant demographic information has on QA systems.

\subsection{Demographics-enhanced Dataset Creation}
We experiment with the following types of modified questions: dimensionless (i.e., no demographic information), ethnicity, gender, names, sexual orientation, gender+ethnicity, gender+sexual orientation, and gender+ethnicity+names. 

The reasoning for each chosen dimension are as follows: dimensionless shows no demographic information, and hence will be used as a baseline to compare how many of the answers change when we add irrelevant demographic information. Ethnicity, sexual orientation, and gender, while not always shown in medical text, are sometimes mentioned when the demographic information is relevant. Hence, we want to see if the models associate any medical conditions with them. We use two genders, but expect that our results will generalize to additional genders.
As for names, these are clearly not medically relevant ever and are rarely shown in medical text. Hence, we choose them to see if there are unexpected differences in answers change. 

Ethnicities include White, Black, African-American, Hispanic, and Asian. Genders include male and female. Names include the 10 names for each ethnicity from the Q-Pain dataset, which originated from the Harvard Dataverse's \textit{Demographic aspects of first names} dataset \cite{DVN/TYJKEZ_2018}. And while “Black” and “African American” are largely synonymous, we want to see if they are different from the models' perspective. Notably, to medically-untrained users, all of these may seem relevant and hence potentially be added to queries when such users request medical assistance.

We follow a similar process as the creators of the Q-Pain dataset and make each context, question, and answer (CQA) as neutral as possible. Given a CQA, such as “A 23-year-old female presents to a psychiatrist...”, we first automatically mask any word that indicates gender (e.g., male, female, he, she, wife, boyfriend): “A 23-year-old [GENDER\_MASK] presents to a psychiatrist...”. 
Then, given a dimension (e.g., gender), we automatically replace each unique masking with their corresponding token replacement (e.g., replacing “[GENDER\_MASK]” with “male”). 

Overall, each of these dimensions and their variations augment each of the 100 vignettes and result in overall 16,700 questions. See Table \ref{Dimensions_example} for examples. And while we only use the English version of the dataset, 
this process can be easily applied to other languages. The data will be publicly available and have an MIT License.

\begin{table*}[t]
\centering
\small
        \setlength{\tabcolsep}{3.0pt}
    \begin{tabular}{l | r r r r |rrrrrrrrrr|}
    \toprule
& \multicolumn{4}{|c|}{\textbf{Names}} & \multicolumn{10}{|c}{\textbf{Gender+Ethnicity+Names}} \\
& \bf\rotatebox{90}{W} &\bf \rotatebox{90}{A-A/B} &\bf \rotatebox{90}{H} &\bf \rotatebox{90}{As} &\bf \rotatebox{90}{M+W} &\bf \rotatebox{90}{M+A-A} &\bf \rotatebox{90}{M+B} &\bf\rotatebox{90}{M+H} &\bf \rotatebox{90}{M+As} &\bf \rotatebox{90}{F+W} &\bf\rotatebox{90}{F+A-A} &\bf \rotatebox{90}{F+B} &\bf \rotatebox{90}{F+H} &\bf \rotatebox{90}{F+As} \\
 \midrule
QAGNN & \textbf{10.5} & \textbf{10.5} & \textbf{12.6} & \textbf{10.5} & \textbf{9.3} & \textbf{14.2} & \textbf{11.5} & \textbf{9.8} & 8.5 & \textbf{9.3} & \textbf{15.0} & \textbf{11.5} & \textbf{12.5} & 7.9 \\

BioLinkBERT & 7.4 & 6.0 & 8.5 & 6.0 & 8.8 & 11.9 & 9.8 & 8.1 & \textbf{9.6} & 8.5 & 11.5 & 10.3 & 10.0 & \textbf{9.0}\\
\bottomrule
    \end{tabular}
    \caption{Percentage of questions with changed answers as compared to a question with no demographic information about the patient. \textit{M}=male; \textit{F}=female; \textit{W}=White; \textit{B}=Black; \textit{A-A}=African-American; \textit{H}=Hispanic; \textit{As}=Asian; \textit{SOr}=sexual orientation. \label{tab:res}}
    \label{names_answers_change_table}
\end{table*}

\begin{table*}[]
\small
\centering
\begin{tabular}{l|clclclclcl}
\toprule
\multicolumn{1}{c|}{\textbf{Model}} & \multicolumn{10}{c}{\textbf{Names}}                                                                                                                                                                                                                                                                                                              \\ \midrule
                                     & \multicolumn{2}{c|}{\textbf{W}}                                   & \multicolumn{2}{c|}{\textbf{B}}                                   & \multicolumn{2}{c|}{\textbf{A-A}}                                  & \multicolumn{2}{c|}{\textbf{H}}                                   & \multicolumn{2}{c}{\textbf{AS}}                                  \\ \midrule
QAGNN                                & \multicolumn{2}{c|}{\textbf{38.6}}                                         & \multicolumn{2}{c|}{\textbf{39.5}}                                         & \multicolumn{2}{c|}{\textbf{39.5}}                                         & \multicolumn{2}{c|}{\textbf{39.3}}                                        & \multicolumn{2}{c}{\textbf{38.5}}                                        \\
BioLinkBert                          & \multicolumn{2}{c|}{38.2}                                        & \multicolumn{2}{c|}{37.3}                                        & \multicolumn{2}{c|}{37.3}                                        & \multicolumn{2}{c|}{37.6}                                         & \multicolumn{2}{c}{37.6}                                         \\ \midrule
                                     & \multicolumn{1}{c}{\textbf{M}} & \multicolumn{1}{c|}{\textbf{F}} & \multicolumn{1}{c}{\textbf{M}} & \multicolumn{1}{c|}{\textbf{F}} & \multicolumn{1}{c}{\textbf{M}} & \multicolumn{1}{c|}{\textbf{F}} & \multicolumn{1}{c}{\textbf{M}} & \multicolumn{1}{c|}{\textbf{F}} & \multicolumn{1}{c}{\textbf{M}} & \multicolumn{1}{c}{\textbf{F}} \\ \midrule
QAGNN                                & \multicolumn{1}{l}{38.6}       & \multicolumn{1}{l|}{38.1}       & \multicolumn{1}{l}{\textbf{39.5}}       & \multicolumn{1}{l|}{\textbf{39.4}}       & \multicolumn{1}{l}{\textbf{39.7}}       & \multicolumn{1}{l|}{\textbf{39.1}}       & \multicolumn{1}{c}{\textbf{39.0}}         & \multicolumn{1}{l|}{\textbf{39.2}}       & \multicolumn{1}{l}{\textbf{39.2}}       & \textbf{37.9}                            \\ 
BioLinkBert                          & \multicolumn{1}{l}{\textbf{39.4}}       & \multicolumn{1}{l|}{\textbf{38.5}}       & \multicolumn{1}{l}{38.6}       & \multicolumn{1}{c|}{37.0}         & \multicolumn{1}{l}{35.1}       & \multicolumn{1}{l|}{38.7}       & \multicolumn{1}{l}{36.8}       & \multicolumn{1}{l|}{37.2}       & \multicolumn{1}{l}{37.3}       & 35.4                            \\ \bottomrule
\end{tabular}
\caption{Accuracy when including names (rows 1 and 2) or names together with gender and ethnicity information (rows 3 and 4) for each model. \textit{W}=White; \textit{B}=Black; \textit{A-A}=African-American; \textit{H}=Hispanic; \textit{As}=Asian;}
\label{accuracy_gender_names_eth}
\end{table*}

\begin{table*}[]
\centering
\small
        \setlength{\tabcolsep}{1.6pt}
    \begin{tabular}{l | r | rr| rrrrr| rrr|rrrrrrrrrr | rrrrrr}
    \toprule
    & \multicolumn{1}{|c|}{\textbf{Random}} 
 & \multicolumn{2}{|c|}{\textbf{Gender}} & \multicolumn{5}{|c|}{\textbf{Ethnicity}} & \multicolumn{3}{|c}{\textbf{SOr}} & \multicolumn{10}{|c}{\textbf{Gender+Ethnicity}}  & \multicolumn{6}{|c}{\textbf{Gender+SOr}} \\ 
 & & \bf \rotatebox{90}{M} &\bf \rotatebox{90}{F} &\bf \rotatebox{90}{W} &\bf \rotatebox{90}{A-A} &\bf \rotatebox{90}{B} &\bf \rotatebox{90}{H} &\bf \rotatebox{90}{As} &\bf \rotatebox{90}{Hetero} &\bf \rotatebox{90}{Bi} &\bf \rotatebox{90}{Homo} & \bf \rotatebox{90}{M+W} &\bf \rotatebox{90}{M+A-A} &\bf \rotatebox{90}{M+B} &\bf\rotatebox{90}{M+H} &\bf \rotatebox{90}{M+As} &\bf \rotatebox{90}{F+W} &\bf\rotatebox{90}{F+A-A} &\bf \rotatebox{90}{F+B} &\bf \rotatebox{90}{F+H} &\bf \rotatebox{90}{F+As} &\bf \rotatebox{90}{M+Hetero} & \bf \rotatebox{90}{M+Bi} & \bf \rotatebox{90}{M+Homo} &\bf \rotatebox{90}{F+Hetero} & \bf \rotatebox{90}{F+Bi} &\bf \rotatebox{90}{F+Homo} \\
 \midrule 
Generic &2 &\textbf{17} &\textbf{16} &6  &\textbf{14} &7 &\textbf{9} &7 &\textbf{9} &\textbf{11} &11 &\textbf{11} &\textbf{11} &\textbf{12} &\textbf{13} &\textbf{8} &\textbf{13} &\textbf{13} &\textbf{13} &\textbf{12} &\textbf{12} &8 &\textbf{10} &6 &\textbf{12} &\textbf{15} &11 \\
Biomedical &2 &6 & 6 & 6 &8 &7 &7 &\textbf{11} &6 &6 &\textbf{14} &6 &7 &5 &7 &6 &8 &8 &9 &8 &8 &\textbf{9} &9 &\textbf{23} &8 &13 &\textbf{23} \\
\bottomrule
    \end{tabular}
    \caption{Percentage of questions with changed answers between the biomedical and generic model as compared to a question with no demographic information about the patient. \textit{M}=male; \textit{F}=female; \textit{W}=White; \textit{B}=Black; \textit{A-A}=African-American; \textit{H}=Hispanic; \textit{As}=Asian; \textit{SOr}=sexual orientation. \label{answers_change_table_generic1}}
\end{table*}

\subsection{Random Change}
\label{baseline}
We use a version of the questions with no demographic information, and, in each prompt's first sentence, replace the word “patient” with “person”. With this we examine the effect of a small but insignificant textual variation on each model. We choose this change over others (e.g., adding random words, irrelevant demographics, or fictitious cities) as this reduces the possibility of models changing their answers due to the context such random words had in the training data (e.g., Africa is more prevalent to the sleeping sickness disease than the US). Moreover, neither “person” nor “patient” reveal information about the human.

\section{Models}
We compare two existing algorithms: QAGNN \cite{qagnn} and BioLinkBert \cite{yasunaga-etal-2022-linkbert}. 
While better models exist for the USMLE dataset, many of them have billions of parameters and we are unable to test them for computational reasons. That being said, BioLinkBert is currently among the state of the art on the USMLE dataset, and QAGNN is the top (and, to the best of our knowledge, only) KG-grounded model. We use existing implementations and models and describe both systems in the following.

\begin{table*}[h]
\small
\centering
        \setlength{\tabcolsep}{1.7pt}
    \begin{tabular}{l | r|  r |  r | rr|rrrrr|rrr|rrrrrrrrrr|rrrrrr  }
    \toprule
    & \bf \textit{O*} & \bf O & \multicolumn{1}{|c|}{\textbf{D}} 
 & \multicolumn{2}{|c|}{\textbf{Gender}} & \multicolumn{5}{|c|}{\textbf{Ethnicity}} & \multicolumn{3}{|c}{\textbf{SOr}} & \multicolumn{10}{|c}{\textbf{Gender+Ethnicity}}  & \multicolumn{6}{|c}{\textbf{Gender+SOr}} \\

 & & & & \bf \rotatebox{90}{M} &\bf \rotatebox{90}{F} &\bf \rotatebox{90}{W} &\bf \rotatebox{90}{A-A} &\bf \rotatebox{90}{B} &\bf \rotatebox{90}{H} &\bf \rotatebox{90}{As} &\bf \rotatebox{90}{Hetero} &\bf \rotatebox{90}{Bi} &\bf \rotatebox{90}{Homo} &  \bf \rotatebox{90}{M+W} &\bf \rotatebox{90}{M+A-A} &\bf \rotatebox{90}{M+B} &\bf\rotatebox{90}{M+H} &\bf \rotatebox{90}{M+As} &\bf \rotatebox{90}{F+W} &\bf\rotatebox{90}{F+A-A} &\bf \rotatebox{90}{F+B} &\bf \rotatebox{90}{F+H} &\bf \rotatebox{90}{F+As} &\bf \rotatebox{90}{M+Hetero} & \bf \rotatebox{90}{M+Bi} & \bf \rotatebox{90}{M+Homo} &\bf \rotatebox{90}{F+Hetero} & \bf \rotatebox{90}{F+Bi} &\bf \rotatebox{90}{F+Homo}\\

 \midrule
1 & \textit{28.9} &26  &25  &29  &27  &27  &26  &26  &27  &27  &28  &27  &26 &27  &27  &27  &26  &25  &27  &27  &27  &26  &24  &27  &27  &28  &25  &26  &25 \\

2 & \textit{\textbf{40}} & \textbf{39} & \textbf{40} & \textbf{40} & \textbf{40} & \textbf{40} & \textbf{38} & \textbf{39} & \textbf{39} & \textbf{36} & \textbf{38} & \textbf{38} & \textbf{37} & \textbf{39} & \textbf{37} & \textbf{41} & \textbf{40} & \textbf{40} & \textbf{40} & \textbf{36} & \textbf{40} & \textbf{41} & \textbf{40} & \textbf{41} & \textbf{41} & \textbf{36} & \textbf{41} & \textbf{40} & \textbf{36} \\

\bottomrule
    \end{tabular}
    \caption{Accuracy (in percentages) of the biomedical and generic models on our demographically enhanced datasets. \textit{M}=male; \textit{F}=female; \textit{W}=White; \textit{B}=Black; \textit{A-A}=African-American; \textit{H}=Hispanic; \textit{As}=Asian; \textit{SOr}=sexual orientation; O*=original test dataset; O=the original, unmodified 100 questions; D=No demographic information; 1=Generic; 2=Biomedical. \label{accuracy_table_generic1}}
\end{table*}

\subsection{QAGNN}
The main component of QAGNN is its KG, which is based on the Disease Database portion of the Unified Medical Language System (UMLS) and DrugBank. The graph contains about 10k nodes and 44k edges, where the embeddings for each node are initialized using the biomedically trained language model SapBERT \cite{sapbert}. SapBERT was trained using the UMLS vocabulary set 2020AA version, which contains biomedical synonyms
from more than 150 controlled vocabularies, such as Gene Ontology and MeSH. QAGNN has 360M parameters.

For each answer choice of a given question, QAGNN first retrieves a subgraph from its KG using entity linking. That is, it finds entity mentions in the question and retrieves any entity in the main KG that appears in any 2-hop paths between pairs of found entities. Then, it concatenates the answer choice and question, followed by encoding using a LM. Next, it connects the encoded representation to the graph as a node. It then performs relevance scoring on each node in the created subgraph by concatenating it to the encoded representation node and calculating the likelihood using a LM. Lastly, using an attention-based graph neural network (GNN) module, it reasons over the graph to get a score for the answer choice. During the training procedure, it optimizes both the LM and its GNN end-to-end using cross-entropy loss. On the MedQA-USMLE dataset, SapBERT-based QAGNN achieves 38\% accuracy.

\subsection{BioLinkBert}
The defining features of BioLinkBert are its pretraining method that incorporates document links and its LM which has similar hyperparameters to PubmedBERT \cite{pubmedbert} and is trained from scratch on the PubMed abstracts PubmedBERT is trained on. BioLinkBert has 340M parameters.

Given a corpus of text, BioLinkBert views it as a graph: it uses Pubmed Parser to extract citation links between documents and views the hyperlinks as edges.
Then, to use the links in its LM pretraining procedure it places two documents which share a link in the same context, in addition to placing two random documents in the same context or a single document (contiguous). Next, it uses two self-supervised objectives. The first, masked language modeling, is common in many of the large LMs such as BERT \cite{devlin-etal-2019-bert}. In the second, document relation prediction, it classifies the link between the two documents as random, linked, or contiguous. On the MedQA-USMLE dataset, the base version of BioLinkBert achieves 40\% accuracy while the large version achieves 44.6\%. Here, we work with the base version because of its lower compute requirements.

\section{Results}
We look at two different effects of providing the model with irrelevant demographic information: 1) the percentage of questions for each model that change and 2) the accuracy change for each model. Note that these are not necessarily correlated: for example, accuracy does not change when initially incorrect answers change to other incorrect answers, or if the same numbers of answers change from incorrect to correct. It is also worth mentioning that \textbf{any change in model’s answers is problematic, as these questions were verified to be independent of demographics.}

\subsection{Changed Answers}
Table \ref{answers_change_table} shows the percentage of questions for each model that change between each dimension's attribute and the dimensionless variation (e.g., between male and genderless).

The first column of Table \ref{answers_change_table}, “Random”, shows the result of our random change (Sec. \ref{baseline}). 
While the other values in the table are larger, and while the words “patient” and “person” may have different connotations for each model based on its training data, this suggests that, to some extent, random noise plays a role in the amount of change each model exhibits. Notably for gender, ethnicity, and sexual orientation, both models change around the same number of answers, except that BioLinkBert has a much higher number for Asian. Additionally, both models have almost double the amount of changed answers for homosexual than bisexual or heterosexual. For gender+ethnicity, QAGNN has an equivalent amount or more than BioLinkBert, though for gender+sexual orientation, BioLinkBert has more than double the amount for homosexuals, with a massive percentage of 23. We also examine the amount of answers for each model for gender, ethnicity, and sexual orientation, that change from being correct to incorrect, from incorrect to correct, and from incorrect to incorrect (Table \ref{correct_to_incorrect_table}). We can see that a model can have an increase in performance (see QAGNN males column which results in a 2\% increase) while having the same number of answers change as a demographics which result in a decrease in performance (see QAGNN White column which result in a 4\% decrease). This implies that accuracy alone is not sufficient to understand the effect irrelevant demographic information has on models' answer, and that further examination of the answers can contribute. For example, we see that adding most ethnicities results in 0 answers changing from incorrect to correct for QAGNN. 

\subsection{Changed Accuracy}
While the reported accuracy on the original test dataset is 38\% for QAGNN and 40\% for BioLinkBert, the accuracy on our 100 randomly selected demographic-independent questions use to construct the vignettes is 40\% for QAGNN and 39\% for BioLinkBert. 
Table \ref{accuracy_table_first} shows our accuracy results for each dimension for each algorithm.

As noted, accuracy change does not always correlate with answer change. For example from Table \ref{answers_change_table}, while both models have about the same number of changed answers for gender, only QAGNN's accuracy for males is affected (increased by 2\%). For ethnicity, both models' accuracy drops, with BioLinkBert's accuracy by 3\% for Asian and QAGNN's accuracy by 4\% for Black. Sexual orientation improves BioLinkBert performance on bisexual and decreases QAGNN's on every variation. Gender+ethnicity decreases QAGNN performance the most (up to 6\%), while gender+sexual orientation improves BioLinkBert's performance on any variation except for homosexual.

\section{Analysis: Names}
Similarly to the above experiments, we also evaluate the effect names have on the two types of models. For names by themselves, for each ethnicity (Black, White, Hispanic, Asian) we use the corresponding 20 names (10 for males and 10 for females). For names+ethnicity+gender, we split the names into their ethnicity and gender. 

Table \ref{names_answers_change_table} and \ref{accuracy_gender_names_eth} show our results: Tables \ref{names_answers_change_table} displays the number of changed answers, while Table \ref{accuracy_gender_names_eth} shows accuracy changes. 
We can see that names alone have a moderate effect on the performance of both models, decreasing the performance in any variation by up to 1.65\%. From our baseline experiment this may be due to random noise. However, by looking at the number of changed answers, we can see that both models have the most change for Hispanics, with QAGNN change of up to 12.6\% and BioLinkBert by up to 8.5\%. Interestingly, QAGNN has the same number of changed answers for White, Black, and Asian, but a different number for Hispanic. More results can be seen in the combination of gender, ethnicity, and names, in which the performance can decrease by up to 3.9\% for BioLinkBert in African American males, and by up to 2.1\% for QAGNN in Asian females. However, the amount of changed answers is up to 15\% in QAGNN for African American females and up to 11.9\% for BioLinkBert in African American males. This implies that even though both models were trained on PubMed data, irrelevant information like names affect them, which is highly problematic.

\section{Medical vs. Generic LMs}
In addition to our main results, we also compare how the performance of a biomedically-trained transformer differs from that of a generic one. In particular, we use the same code for the BioLinkBert QA system, but instead of using the medically-trained base (trained from scratch on PubMed abstracts), we use a transformer which is trained on generic English text. 

Similar to our analysis between QAGNN and BioLinkBert above, our analysis between the biomedical and generic models can be split into the amount of answers and accuracy that changes when the dimensions change. From Table \ref{answers_change_table_generic1} it is visible that the generic transformer has more than double the amount of answers change for each gender. It also has an equivalent amount or more for almost any ethnicity, except for Asians. Notably, for sexual orientation, the generic transformer has almost double the amount of answers change for bisexuals, while the biomedical transformer has more for homosexuals. The generic transformer has significantly larger values than the biomedical transformer in any gender+ethnicity combination, while for gender+sexual orientation, the biomedical system has significantly larger values for homosexuals. From Table \ref{accuracy_table_generic1} it is clear that BioLinkBert significantly outperforms its generic LM variation. From the change in accuracy we can see that, while the biomedical transformer's accuracy increases when gender is removed (“no info”), the generic transformer's accuracy decreases. We can also see that the biomedical transformer's accuracy changes more for ethnicity and sexual orientation, while the generic model changes more for gender.

\section{Future Work}
Finally, we discuss three potential approaches to alleviate the aforementioned effects, including model architectures, data, and regularization. 

\paragraph{Model Architecture}
While both the KG-grounded LM and the text-based one are susceptible to irrelevant demographic information, our initial assumption that the KG-based LM would be less susceptible still holds. In particular, KGs are a condensed representation of knowledge, which rarely holds such irrelevant information. Hence, models that use such representations have a significant potential to be less affected. That being said, a potential reason that the tested KG-based LM is still susceptible may be due to the fact that it grounds the text using the KG, and does not only uses the KG. Hence, the demographically irrelevant information may still leak into the final representation, which the model uses to answer the question. 

\paragraph{Data}
Generally, large LMs are trained using a massive corpus. This is problematic as it is almost impossible to ensure that every piece of data is demographically independent. To try to alleviate this issue, we select biomedical models that are trained only on biomedical data, which often does not contain demographically irrelevant information. However, we still find that these models are susceptible to such information. Hence, future work should examine methods to reduce such issues in the training data, especially for models intended to critical settings.

\paragraph{Regularization}
While developing models with different architectures or ensuring that every piece of data is demographically independent is time consuming, a potentially simple method to alleviate such problem is to regularize the input itself. For example, by masking demographically-significant words. And while relatively simple to implement (e.g., using keywords search), in medicine it is sometimes the case where such demographically-significant words are in fact significant. Hence, simple masking might reduce bias, but will also reduce performance. Future work should examine potential masking approaches that consider times where such words are actually needed. 

\section{Conclusion}
We examine the effect of irrelevant demographic information on purely text-based and KG-grounded biomedical QA systems as well as a generic QA system, using a subset of the USMLE questions whose answers do not depend on the patient's demographics. Our results show that irrelevant demographic information results in changed answers for all systems. We also find that, while all systems are affected by irrelevant demographic information, they differ with regards to how different types of demographic information influence them. These results provide evidence that more work is needed in order to ensure fair treatment of all patients by biomedical QA systems.

\section*{Limitations}
As expert annotation is expensive, we only use 100 unique vignettes to create the $16,700$ questions. However, this is almost twice as many as other published datasets, such as \citet{qpain}. Additionally, we  only analyze one KG-grounded and one purely text-based system. While our main point, that there are problems one should be aware of, can be made based on experiments with two models, evaluating more systems can potentially lead to more fine-grained insights.

\section*{Ethics Statement}
The main reason for this paper is to point out potential problems regarding fair treatment of all patients by biomedical QA systems. Future work should improve existing biomedical QA systems to ensure equal and just patient care. Moreover, such systems can be problematic for both patients and health experts. For example, a patient could follow the recommendations of such a QA model at home without expert supervision and a system could recommend an incorrect treatment because of their name, or physicians could use such systems to improve their quality of care, but the system could cloud their judgment and direct them to an incorrect answer.

\section*{Acknowledgments}
We thank Dr. Peter Pressman for his help with reviewing the data and the reviewers for their feedback. The authors acknowledge financial support from NIH grants OT2TR003422 and R01LM013400.

\bibliography{anthology,custom}
\bibliographystyle{acl_natbib}
\end{document}